\documentclass[letterpaper, 10 pt, conference]{ieeeconf}  
\usepackage{xcolor}

\IEEEoverridecommandlockouts                              

\usepackage{booktabs}
\usepackage{multirow}
\usepackage{graphicx}
\usepackage{capt-of}
\usepackage{amsmath} 
\usepackage{amssymb}  
\let\labelindent\relax
\usepackage{enumitem}
\usepackage{hyperref}
\title{\LARGE \bf
Robotizing Human Videos with Physically Consistent Interactions
}

\author{\authorblockN{Ching-Lam Cheng, Shengfeng He \authorrefmark{2}, and Bin Zhu \textsuperscript{\authorrefmark{2}}}
\authorblockA{Singapore Management University \\
cl.cheng.2024@phdcs.smu.edu.sg, shengfenghe@smu.edu.sg, binzhu@smu.edu.sg \\ 
\url{https://jacquescheng.github.io/robotize-consistent-interaction/}
\thanks{\authorrefmark{2}Corresponding author and project lead.}}
}

\begin{document}

\IEEEaftertitletext{%
\vspace{-10mm}
\begin{center}
    \includegraphics[width=\textwidth]{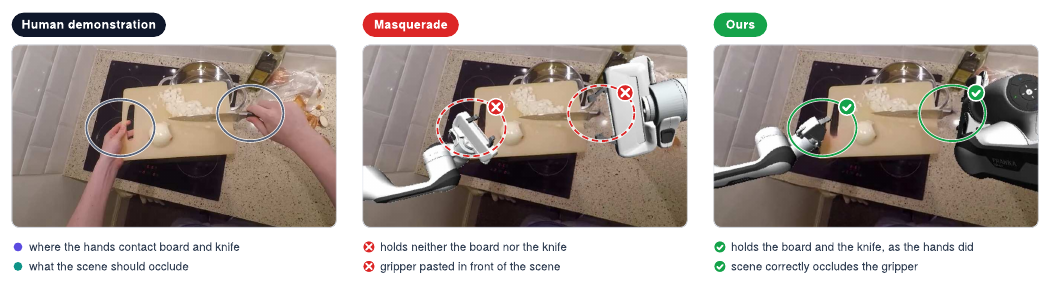}\vspace{-4mm}
    \captionof{figure}{We propose an interaction-aware, depth-consistent human-to-robot video-editing pipeline that preserves both grasp geometry and robot--object visibility. Existing editing methods such as Masquerade~\cite{lepert2025masquerade} can produce grippers that miss the demonstrated contacts and violate scene occlusions, whereas our method better preserves the physical interaction in the original human demonstration.}
    \label{fig:teaser}
\end{center}
}
\maketitle

\thispagestyle{empty}
\pagestyle{empty}


\newcommand{\todo}[1]{\textcolor{red}{[TODO: #1]}}


\begin{abstract}
Human videos offer scalable manipulation data, but the embodiment gap between human hands and robot manipulators limits their direct use. Existing video-editing methods replace hands with rendered robots, yet inaccurate interaction reconstruction and compositing can produce inconsistent grasps and implausible robot–object occlusions. We address these failures from two complementary physical aspects: interaction geometry and scene visibility. First, an interaction-aware contact reconstruction module combines hand–object segmentation with mesh-level contact prediction to recover dense 3D contacts, then converts them into temporally stabilized grasps for parallel-jaw grippers. Second, a depth-aware compositing module uses scene and robot depth to enforce physically consistent robot–object occlusions. The resulting videos preserve the interaction structure of human demonstrations in a robot-compatible form and are co-trained with robot demonstrations. Using identical human videos and robot data, we compare against robot-only training and the original Masquerade pipeline. Across four RoboTwin tasks and two Diffusion Policy visual encoders, our method achieves the highest average success rates, with especially strong gains under out-of-distribution scene variation.
 Real-world deployment further shows that the proposed co-training approach improves robustness to visual distractors when the task geometry is observable, while performance on depth-sensitive grasps remains limited by the single-camera setup.

\end{abstract}

\section{Introduction}
The advancement in robot learning has increasingly benefited from large and diverse demonstration datasets~\cite{pmlr-v305-black25a, ye2026worldactionmodelszeroshot}. However, collecting robot demonstrations at scale remains fundamentally expensive because each trajectory requires physical hardware, environment interaction, and often human teleoperation or supervision~\cite{o2024open, khazatsky2024droid}. Consequently, robot datasets remain far smaller and less diverse than the image, text, and video corpora that have driven progress in large language models~\cite{achiam2023gpt, yang2025qwen3} and video generation models~\cite{wan2025wan}. This data bottleneck limits the diversity of interactions available during training and, in turn, the ability of manipulation policies to generalize across objects, tasks, and environments.

Human videos offer a compelling alternative source of manipulation experience. Large-scale egocentric video collections~\cite{Damen2018EPICKITCHENS, punamiya2026egoverse, perrett2025hd} capture everyday interactions across diverse objects, scenes, and behaviors at a scale that is difficult to reproduce with physical robots. In principle, these videos could provide robot policies with substantially broader manipulation experience. In practice, however, human demonstrations differ fundamentally from robot demonstrations~\cite{yang2025egovla, zheng2026egoscale}. Human hands differ from robot manipulators in appearance, kinematics, and contact mechanisms, while human videos typically lack robot-compatible action labels. This embodiment gap prevents conventional robot learning pipelines from directly treating human videos as robot demonstrations.

Recent work has begun to bridge this gap by converting human demonstrations into robot-compatible visual data~\cite{yang2025egovla, zheng2026egoscale}. In particular, Phantom~\cite{lepert2025phantom} and Masquerade~\cite{lepert2025masquerade} introduce a video-editing paradigm that reconstructs human hand motion from egocentric video, removes the human hand through inpainting, and renders a robot manipulator along the recovered trajectory, as shown in the middle of Fig.~\ref{fig:teaser}. The resulting robotized videos reduce the visual embodiment discrepancy and can be incorporated into robot policy training. This offers a promising route for scaling robot learning with abundant human video.

However, successful human-to-robot conversion requires more than replacing the visual embodiment. A manipulation demonstration is informative because it encodes a physically coherent interaction between the agent, object, and scene. If the conversion process alters this interaction structure, the edited video may look robot-like while providing inconsistent supervision. This issue is particularly important because human-to-robot editing is a multi-stage process in which errors from hand reconstruction, grasp retargeting, and scene compositing can accumulate. For example, retargeting a robot gripper from hand pose or motion alone does not explicitly preserve where the human contacts the object, and can therefore produce a gripper pose that is kinematically plausible but inconsistent with the demonstrated grasp. Likewise, naively overlaying a rendered robot onto the edited scene ignores scene geometry and can place gripper pixels in front of object regions that should physically occlude them. Such artifacts create disagreement between robot motion, object interaction, and visual evidence, weakening the value of the converted video as policy supervision. We therefore argue that the key challenge is not only to reduce the appearance gap between humans and robots, but also to preserve the physical consistency of the demonstrated interaction.

To this end, we propose an interaction-faithful human-to-robot video-editing pipeline that addresses physical consistency from two complementary aspects: \emph{interaction geometry} and \emph{scene visibility}. First, we explicitly reconstruct hand--object contact rather than deriving the robot grasp from hand motion alone. Our interaction-aware contact reconstruction combines hand--object segmentation with mesh-level contact prediction to recover dense 3D contact regions on the human hand. We then reduce these contacts to a temporally stabilized thumb--opposing-finger contact pair that can be naturally mapped to a parallel-jaw gripper. Surface-normal constraints further favor contact pairs that correspond to geometrically plausible opposing grasp surfaces. This formulation grounds robot retargeting in the demonstrated hand--object interaction itself, rather than only in the reconstructed hand configuration. Second, we introduce depth-aware robot--scene compositing to preserve physically correct visibility. By comparing the predicted depth of the hand-inpainted scene with the rendered robot depth, we identify gripper regions that should lie behind the manipulated object and remove them from the final composite. Together, as shown in the right of Fig.~\ref{fig:teaser}, these components preserve both where the robot should interact with the object and how that interaction should appear in the scene.

We evaluate the resulting robotized demonstrations by co-training manipulation policies with real robot demonstrations. Using the same source human videos and robot training data, we compare our pipeline against robot-only training and the original Masquerade editing pipeline. Experiments span four RoboTwin manipulation tasks and two visual encoders within Diffusion Policy. Our edited demonstrations consistently improve policy performance, achieving the highest average success rates across settings. The gains are particularly pronounced under out-of-distribution scene variation, suggesting that physically consistent human-video supervision provides useful interaction cues beyond the visual conditions observed in the robot training data.
We further validate the approach on a physical robot, where it improves robustness to visual distractors for tasks with sufficiently observable interaction geometry. Results on a depth-sensitive grasp also reveal the limitations imposed by ambiguous geometric cues in a single-camera setup.

Our main contributions are:
\begin{itemize}[leftmargin=*, nosep]
\item We identify \emph{interaction consistency} as a key requirement for converting human videos into reliable robot-learning supervision.
\item We introduce interaction-aware contact reconstruction that recovers dense 3D hand--object contacts and retargets them into stable, geometrically plausible parallel-jaw grasps.
\item We propose depth-aware compositing that preserves physically consistent robot--object occlusions in edited videos.
\item We show across four RoboTwin tasks, two Diffusion Policy visual encoders, and real-robot experiments that our edited demonstrations improve average performance and robustness to out-of-distribution scene variation.
\end{itemize}

\section{Related Work}
\label{sec:related_work}

\noindent\textbf{Learning from egocentric human video.}
Egocentric video offers a scalable source of human manipulation experience from a viewpoint closely aligned with robot observations. Large-scale datasets such as EPIC-KITCHENS~\cite{Damen2018EPICKITCHENS}, HD-EPIC~\cite{perrett2025hd}, and EgoVerse~\cite{punamiya2026egoverse} capture diverse everyday interactions across objects, tasks, and environments. Despite this scale and diversity, directly using such videos for robot learning remains difficult because they depict human hands rather than robot manipulators and generally lack robot-compatible action labels, creating both embodiment and action-space gaps. 
Several works address these gaps by incorporating human demonstrations directly into robot policy learning. EgoMimic~\cite{kareer2024egomimicscalingimitationlearning} tracks human hand motion and aligns egocentric human demonstrations with robot data for joint imitation learning. EgoVLA~\cite{yang2025egovla} leverages egocentric human video as supervision for vision--language--action models, while EgoScale~\cite{zheng2026egoscale} studies how diverse egocentric data can scale dexterous manipulation learning. These approaches demonstrate that human videos can complement expensive robot-collected trajectories, but require learning mechanisms that explicitly accommodate the mismatch between human and robot embodiments.

\begin{figure*}[t]
\centering
\includegraphics[width=\linewidth]{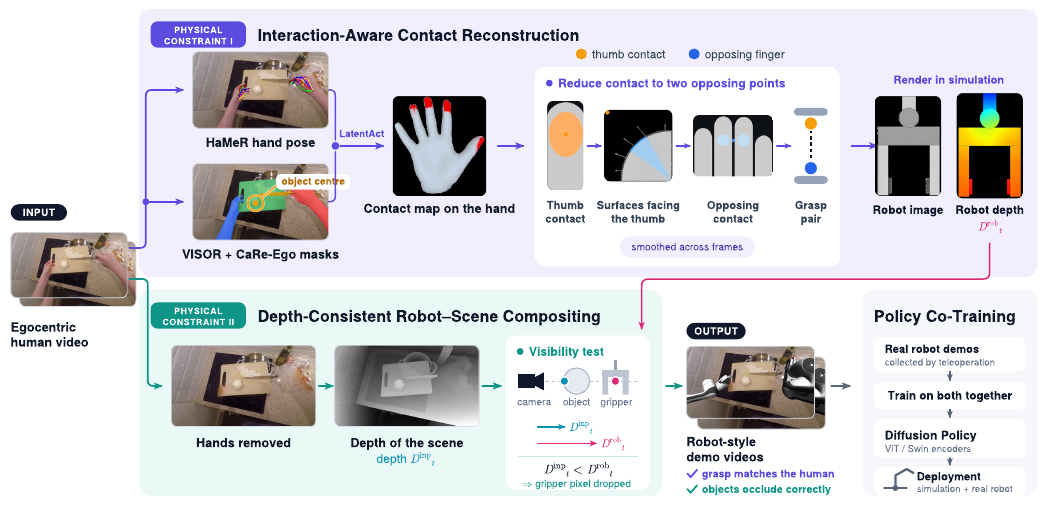}\vspace{-2mm}
\caption{Overview of our framework. (a) Interaction-aware contact reconstruction combines hand--object segmentation and mesh-level contact prediction to recover temporally stable opposing contacts for robot grasp retargeting. (b) Depth-consistent compositing compares scene and rendered-robot depth to enforce physically correct robot--object occlusions. The resulting robotized videos are co-trained with real robot demonstrations for policy learning.
}
\label{fig:method}
\end{figure*}

\noindent\textbf{Human-to-robot video editing.}
A complementary line of work addresses the embodiment gap at the data level by converting human videos into robotized demonstrations before policy training. Phantom~\cite{lepert2025phantom} explores learning robot manipulation from human videos without requiring robot demonstrations. Masquerade~\cite{lepert2025masquerade} reconstructs human hand motion, removes the human hand through video inpainting, and renders a target robot along the recovered trajectory. The resulting videos visually resemble robot demonstrations and can be co-trained with conventional robot data, providing a direct way to reuse human manipulation experience within standard robot-learning pipelines.
Our work follows this data-editing and co-training paradigm, but focuses on a complementary problem: whether the converted video preserves the physical interaction encoded in the original demonstration. Existing editing pipelines primarily bridge the visual embodiment gap, while errors in grasp retargeting and scene compositing can still produce robotized demonstrations with inconsistent interaction geometry or visibility. We address these limitations from two complementary aspects. First, we incorporate explicit hand--object contact geometry when retargeting human grasps to robot grippers. Second, we use depth-based occlusion reasoning when compositing the rendered robot into the scene. Together, these components improve both the physical interaction and its visual consistency, producing more faithful robot-learning supervision from the same egocentric source videos.

\section{Methodology} 
\subsection{Problem Formulation and Overview}
\label{sec:problem_setup}

We consider the problem of converting egocentric human manipulation videos into robotized demonstrations for policy co-training with conventional robot demonstrations. Given an RGB video sequence $\mathcal{V}=\{I_t\}_{t=1}^{T}$ and a target robot equipped with parallel-jaw grippers, our goal is to recover robot-compatible motion and grasp information and generate an edited video in which the robot replaces the human hands. The conversion should preserve the contact geometry of the human demonstration and the visibility relationships between the rendered gripper and the surrounding scene.

Following the data-editing framework of Masquerade~\cite{lepert2025masquerade}, we use reconstructed 3D hand motion as the basis for robot retargeting. At frame $t$, the hand states are denoted by $\mathcal{H}_t=\{\mathbf{H}_t^{l},\mathbf{H}_t^{r}\}$, where $\mathbf{H}_t^{h}$ specifies the pose and articulation of hand $h\in\{l,r\}$ through a reconstructed MANO mesh. To obtain interaction cues, we use the sparse VISOR annotations~\cite{VISOR2022} associated with EPIC-KITCHENS~\cite{Damen2018EPICKITCHENS} and augment them with CaRe-Ego~\cite{su_care-ego_2026} to obtain dense hand and interacting-object masks. We denote these masks by $M_t^h$ and $M_t^o$, respectively, with $M_t^o$ referring to the object associated with the hand under consideration. The subsequent contact reconstruction is applied to each hand, and we omit the hand superscript where the context is clear.

As shown in Fig.~\ref{fig:method}, our method addresses two aspects of this conversion. First, interaction-aware contact reconstruction (Sec.~\ref{sec:contact_reconstruction}) uses the object-mask centroid and reconstructed hand geometry to estimate mesh-level contact. The estimated contact region is reduced to a temporally stabilized thumb--opposing-finger contact pair and an opposing-finger identity for retargeting. Second, depth-based robot occlusion estimation (Sec.~\ref{sec:depth_occlusion}) compares the depth of the hand-inpainted scene with that of the rendered robot to determine gripper visibility during compositing.

The output is a robotized demonstration
\begin{equation}
    \mathcal{D}_{R}=\{(I_t^{\mathrm{edit}},\mathcal{R}_t)\}_{t=1}^{T},
\end{equation}
where $I_t^{\mathrm{edit}}$ is the composited robotized frame and $\mathcal{R}_t$ denotes the corresponding retargeted robot state, including its gripper configuration. These demonstrations augment the original robot dataset for policy co-training. Our focus is on improving demonstration generation; the downstream policy architecture and training configuration are held fixed within each experimental comparison.

\subsection{Interaction-Aware Contact Reconstruction}
\label{sec:contact_reconstruction}

Accurate hand--object contact is essential for transferring human manipulation to a robot, as it determines not only when the robot gripper should be activated, but also which regions of the human hand provide the most relevant geometric constraints for retargeting. To recover this information from egocentric video, we construct a dense 3D contact representation by combining contact-aware hand--object segmentation with mesh-level contact estimation. As illustrated in Fig.~\ref{fig:method}, our contact reconstruction pipeline consists of three main stages. First, we obtain dense hand and interacting-object masks by augmenting VISOR~\cite{VISOR2022} with CaRe-Ego~\cite{su_care-ego_2026}. Second, the centroid of the segmented interacting-object mask is used as an image-space contact cue for LatentAct~\cite{Prakash2025LatentAct}, which predicts a dense contact map over the reconstructed MANO hand mesh. Finally, the predicted contact region is partitioned into thumb and non-thumb regions and reduced to two representative 3D contact points, which provide a compact grasp representation compatible with a parallel-jaw robot gripper. Temporal stabilization and surface-normal filtering are further applied to improve the consistency of the extracted contacts across frames.

\paragraph{Contact-Aware Hand--Object Segmentation.}
VISOR~\cite{VISOR2022} provides pixel-level annotations for hands and active objects; however, the annotations are temporally sparse and do not directly provide dense 3D hand contact. We augment the VISOR annotations using CaRe-Ego~\cite{su_care-ego_2026}, which explicitly models the contact relationship between hands and interacting objects. In particular, CaRe-Ego employs hand features to enhance the representation of interacting objects and decouples contact-centric object features associated with different hands. We use the resulting segmentation to obtain dense hand and interacting-object masks for each frame, denoted by $M_t^h$ and $M_t^o$, respectively.

For each frame, we compute the centroid of the interacting-object mask as
\begin{equation}
    \mathbf{q}_t^o
    =
    \frac{1}{|M_t^o|}
    \sum_{\mathbf{x}\in M_t^o}\mathbf{x},
\end{equation}
where $\mathbf{x}\in\mathbb{R}^2$ denotes a foreground pixel in the object mask. This centroid provides a compact image-space representation of the object location associated with the current interaction. Rather than treating the entire object mask as a contact region, we use $\mathbf{q}_t^o$ as the contact input to LatentAct. The centroid is first expressed in the image coordinate system expected by LatentAct and is then combined with the reconstructed MANO hand pose and camera parameters. Thus, the model receives the estimated hand configuration together with the image-space location of the interacting object, allowing it to infer which regions of the hand mesh are most likely to be in contact with the object.

Following LatentAct~\cite{Prakash2025LatentAct}, the model predicts a contact map over the reconstructed MANO hand mesh. Let $\mathbf{v}_{t,i}\in\mathbb{R}^3$ denote the $i$-th MANO vertex and let $\Pi(\mathbf{v}_{t,i})$ denote its projection onto the image plane. LatentAct uses the projected hand geometry and the object-centroid contact input $\mathbf{q}_t^o$ to produce a per-vertex contact prediction,
\begin{equation}
    \mathbf{c}_t
    =
    \operatorname{LatentAct}
    \left(
    \{\mathbf{v}_{t,i}\}_{i=1}^{778},
    \{\Pi(\mathbf{v}_{t,i})\}_{i=1}^{778},
    \mathbf{q}_t^o
    \right),
\end{equation}
where $\mathbf{c}_t\in[0,1]^{778}$ is the predicted contact map. Each value $c_{t,i}$ represents the estimated likelihood that MANO vertex $i$ participates in the hand--object interaction. We convert this continuous prediction into a binary contact state using a threshold $\delta$,
\begin{equation}
    \hat{c}_{t,i}
    =
    \mathbb{I}[c_{t,i}>\delta],
\end{equation}
and construct the corresponding dense 3D contact mesh as
\begin{equation}
    \mathcal{P}_t
    =
    \{\mathbf{v}_{t,i}\mid \hat{c}_{t,i}=1\}.
\end{equation}

This procedure uses the object-mask centroid as the spatial cue that anchors LatentAct's contact prediction to the current interaction. The centroid does not itself specify the contacting finger or a precise 3D point; instead, LatentAct propagates this image-space cue through the reconstructed hand geometry to estimate a distributed contact region on the MANO surface. Retaining this dense contact mesh preserves information about the spatial extent and finger configuration of the human grasp, in contrast to estimating a single contact point directly from the image.

\paragraph{Gripper Contact Point Extraction.}
The human hand may establish contact with an object using multiple fingers and a large surface area, whereas the target parallel-jaw gripper requires only two representative contact points. We therefore reduce $\mathcal{P}_t$ into a thumb-side contact point and an opposing-finger contact point.

We first partition the contact mesh according to the MANO vertex-to-finger assignment into the thumb contact set $\mathcal{P}_t^{\mathrm{th}}$ and the remaining non-thumb contact set $\mathcal{P}_t^{\mathrm{nt}}$. When thumb contact is detected, the thumb-side contact point is computed as the centroid of the corresponding contact vertices,
\begin{equation}
    \mathbf{p}_t^{\mathrm{th}}
    =
    \frac{1}{|\mathcal{P}_t^{\mathrm{th}}|}
    \sum_{\mathbf{v}\in\mathcal{P}_t^{\mathrm{th}}}\mathbf{v}.
\end{equation}
If no valid thumb contact is detected, we instead use the reconstructed thumb fingertip as a fallback.

Not all non-thumb contact vertices form a meaningful opposing surface for a parallel-jaw grasp. We therefore additionally consider the surface normal $\mathbf{n}_{t,i}$ of each non-thumb contact vertex. For a candidate vertex $\mathbf{v}_{t,i}$, we compute the unit direction toward the thumb contact as $\mathbf{d}_{t,i}=(\mathbf{p}_t^{\mathrm{th}}-\mathbf{v}_{t,i})/\|\mathbf{p}_t^{\mathrm{th}}-\mathbf{v}_{t,i}\|_2$. We retain only vertices whose surface normals sufficiently face the thumb,
\begin{equation}
    \widetilde{\mathcal{P}}_t^{\mathrm{nt}}
    =
    \left\{
    \mathbf{v}_{t,i}\in\mathcal{P}_t^{\mathrm{nt}}
    \mid
    \mathbf{n}_{t,i}^{\top}\mathbf{d}_{t,i}>\tau
    \right\},
    \qquad \tau=0.3.
\end{equation}
This filtering suppresses contact vertices located on surfaces whose orientation is inconsistent with an opposing grasp. If no vertex satisfies the normal constraint, we revert to the complete non-thumb contact set $\mathcal{P}_t^{\mathrm{nt}}$.

The opposing contact point is then obtained from the centroid of the filtered vertices,
\begin{equation}
    \mathbf{p}_t^{\mathrm{nt}}
    =
    \frac{1}{|\widetilde{\mathcal{P}}_t^{\mathrm{nt}}|}
    \sum_{\mathbf{v}\in\widetilde{\mathcal{P}}_t^{\mathrm{nt}}}\mathbf{v}.
\end{equation}
When no non-thumb contact is available, we use the middle-finger fingertip as the fallback contact point.

\paragraph{Temporal Contact Stabilization.}
Frame-wise contact estimation can fluctuate due to segmentation errors, hand reconstruction noise, and small changes in the estimated object-mask centroid or LatentAct contact map. Such fluctuations are undesirable when the extracted contacts are subsequently converted into robot trajectories. We therefore temporally smooth each valid contact point with its corresponding contact from the previous valid frame. Given the current estimate $\mathbf{p}_t$ and previous valid contact $\mathbf{p}_{t^-}$, the stabilized contact is computed as
\begin{equation}
    \bar{\mathbf{p}}_t
    =
    \lambda\mathbf{p}_{t^-}
    +(1-\lambda)\mathbf{p}_t,
    \qquad \lambda=0.3.
\end{equation}
We apply the same procedure independently to the thumb and non-thumb contacts.

\paragraph{Grip Finger Selection.}
Finally, we determine which non-thumb human finger should be mapped to the opposing side of the robot gripper. Since the centroid $\mathbf{p}_t^{\mathrm{nt}}$ does not necessarily coincide with a MANO mesh vertex, we identify its nearest vertex within the filtered contact set,
\begin{equation}
    i_t^*
    =
    \arg\min_{i:\mathbf{v}_{t,i}\in
    \widetilde{\mathcal{P}}_t^{\mathrm{nt}}}
    \|\mathbf{v}_{t,i}-\mathbf{p}_t^{\mathrm{nt}}\|_2.
\end{equation}
Using the predefined MANO vertex-to-finger mapping, the finger associated with $i_t^*$ is selected as the active opposing finger. If no valid non-thumb contact is detected, we default to the middle finger. The final human grasp is therefore represented by the pair $(\bar{\mathbf{p}}_t^{\mathrm{th}},\bar{\mathbf{p}}_t^{\mathrm{nt}})$ together with the identity of the opposing finger. This compact representation preserves the dominant human contact geometry inferred by LatentAct while being directly compatible with subsequent retargeting to a two-finger parallel-jaw gripper.

\subsection{Depth-Consistent Robot–Scene Compositing}
\label{sec:depth_occlusion}

After retargeting the recovered human hand trajectory to the robot embodiment, we render the robot into the egocentric scene to generate robotized demonstrations. A key challenge in this step is handling occlusion between the rendered robot gripper and the real manipulated object. Directly overlaying the rendered robot on the raw frame can produce physically implausible images, because object regions that should appear in front of the robot may be incorrectly covered by the rendered gripper.

To address this issue, we estimate occlusion using the inpainted video frame rather than the raw egocentric frame. Following the visual editing procedure of Masquerade~\cite{lepert2025masquerade}, we remove the segmented human hands using E2FGVI video inpainting~\cite{Li_2022_CVPR_E2FGVI}. Let $\widetilde{I}_t$ denote the resulting hand-inpainted frame. As illustrated in Fig.~\ref{fig:method}, we predict a depth map $D_t^{\mathrm{inp}}$ from $\widetilde{I}_t$, so that the depth estimate corresponds to the scene after human-hand removal. We also render the retargeted robot in simulation and obtain its depth map $D_t^{\mathrm{rob}}$ together with a robot-body mask $M_t^{\mathrm{rob}}$ excluding the gripper, and a separate gripper mask $M_t^{\mathrm{grip}}$.

For each pixel $\mathbf{x}$ covered by the rendered gripper, we compare the inpainted-scene depth with the rendered robot depth. If $D_t^{\mathrm{inp}}(\mathbf{x}) < D_t^{\mathrm{rob}}(\mathbf{x})$, the scene point is closer to the camera than the rendered gripper and should therefore occlude it. The final overlay mask is computed as
\begin{equation}
    M_t^{\mathrm{overlay}}(\mathbf{x})
    =
    M_t^{\mathrm{rob}}(\mathbf{x})
    \lor
    \left(
    M_t^{\mathrm{grip}}(\mathbf{x})
    \land
    D_t^{\mathrm{inp}}(\mathbf{x}) \geq D_t^{\mathrm{rob}}(\mathbf{x})
    \right).
\end{equation}
This depth comparison removes gripper pixels that lie behind the inpainted scene while keeping the visible robot body in the overlay. Using the inpainted frame is important because the raw frame still contains the human hands, whose depth would incorrectly compete with the rendered robot and introduce spurious occlusions after the hands have been removed.

\subsection{Robotized Demonstrations for Policy Co-Training}
\label{sec:policy_cotraining}

We follow the policy-learning procedure of Masquerade~\cite{lepert2025masquerade}. Let
\begin{equation}
    \mathcal{D}_{\mathrm{edit}}
    =\{(I_t^{\mathrm{edit}},\mathbf{p}_{t:t+H}^{\mathrm{2D}})\},
    \qquad
    \mathcal{D}_{\mathrm{robot}}
    =\{(\mathbf{o}_t^r,\mathbf{a}_t^r)\},
\end{equation}
denote the robotized human-video dataset and task-specific robot demonstration dataset, respectively. The visual encoder is first pre-trained on $\mathcal{D}_{\mathrm{edit}}$ to predict future 2D robot waypoints $\mathbf{p}_{t:t+H}^{\mathrm{2D}}$. During policy learning, this auxiliary objective is retained while the encoder and policy head are optimized on the robot observations $\mathbf{o}_t^r$ and actions $\mathbf{a}_t^r$. The overall objective is
\begin{equation}
    \mathcal{L}
    =\mathcal{L}_{\mathrm{2D}}(\mathcal{D}_{\mathrm{edit}})
    +\lambda_{\mathrm{policy}}
    \mathcal{L}_{\mathrm{policy}}(\mathcal{D}_{\mathrm{robot}}).
\end{equation}
Thus, the edited videos provide visual representation supervision, whereas the robot demonstrations provide robot-action supervision. We keep this training procedure fixed and modify only the human-to-robot editing pipeline, allowing the experiments to isolate the effect of improving the quality of the robotized demonstrations.

\begin{table*}[t]
    \centering
    \caption{Task success rates (\%) on RoboTwin over 100 evaluation rollouts. All policies are trained in the ID scene. ID and OOD denote evaluation without and with visual distractors, respectively. Average values are computed across the four tasks. Bold indicates the best result for each policy backbone and evaluation setting.}
    \label{tab:robotwin_results}
    \small
    \setlength{\tabcolsep}{3.2pt}
    \begin{tabular}{ll*{5}{cc}}
        \toprule
        \multirow{2}{*}{Policy} & \multirow{2}{*}{Training data}
        & \multicolumn{2}{c}{Adjust Bottle}
        & \multicolumn{2}{c}{Grab Roller}
        & \multicolumn{2}{c}{Place Burger Fries}
        & \multicolumn{2}{c}{Shake Bottle}
        & \multicolumn{2}{c}{Average} \\
        \cmidrule(lr){3-4} \cmidrule(lr){5-6} \cmidrule(lr){7-8}
        \cmidrule(lr){9-10} \cmidrule(lr){11-12}
        & & ID & OOD & ID & OOD & ID & OOD & ID & OOD & ID & OOD \\
        \midrule
        \multirow{3}{*}{DP-ViT}
            & Baseline   & 33 & 15 & 27 & 1  & 40 & 21 & \textbf{69} & 29 & 42.25 & 16.50 \\
            & Masquerade & 76 & 37 & \textbf{85} & 12 & 37 & 23 & 64 & 22 & 65.50 & 23.50 \\
            & Ours       & \textbf{82} & \textbf{70} & 75 & \textbf{33} & \textbf{46} & \textbf{43} & \textbf{69} & \textbf{37} & \textbf{68.00} & \textbf{45.75} \\
        \midrule
        \multirow{3}{*}{DP-Swin}
            & Baseline   & 73 & 30 & 80 & 22 & 39 & 10 & 84 & 10 & 69.00 & 18.00 \\
            & Masquerade & 94 & 57 & 94 & 43 & 69 & 39 & 84 & \textbf{36} & 85.25 & 43.75 \\
            & Ours       & \textbf{97} & \textbf{84} & \textbf{95} & \textbf{74} & \textbf{72} & \textbf{58} & \textbf{85} & 21 & \textbf{87.25} & \textbf{59.25} \\
        \bottomrule
    \end{tabular}
\end{table*}

\section{Experiment}
\subsection{Experimental Setup}
\label{sec:exp_setup}

\paragraph{Comparison Methods.}
We compare three training settings to separate the benefit of adding human-video-derived demonstrations from the benefit of improving their conversion quality.

\textbf{Baseline (w/o Co-training).}
The policy is trained only on the original robot demonstrations. This represents the standard robot-learning setting and provides a baseline without human-video co-training.

\textbf{Masquerade.}
Following Masquerade~\cite{lepert2025masquerade}, we augment the robot data with egocentric human videos converted to the target robot embodiment using its original data-editing pipeline. The robot and converted human demonstrations are then used jointly for policy training.

\textbf{Ours.}
We use the same human videos and robot demonstrations as Masquerade, but process the videos with our interaction-aware pipeline. Our conversion incorporates contact-aware hand--object segmentation, 3D contact reconstruction, contact-guided grasp retargeting, and depth-based robot--scene compositing.

Both simulation and real-world experiments use 50 robot demonstrations per task. Masquerade and our method also use the same source human videos. We hold the policy architecture and training configuration fixed within each comparison, so differences between these two settings primarily reflect the human-to-robot conversion pipeline.

\paragraph{Evaluation Scenes.}
For each task, the in-distribution (ID) scene contains only task-relevant objects, whereas the out-of-distribution (OOD) scene adds irrelevant objects as visual distractors. All policies are trained using ID-scene demonstrations and evaluated in both scenes. This protocol measures performance under the training distribution and robustness to changes in visual context.

\begin{figure}[t]
    \centering
    \includegraphics[width=\columnwidth]{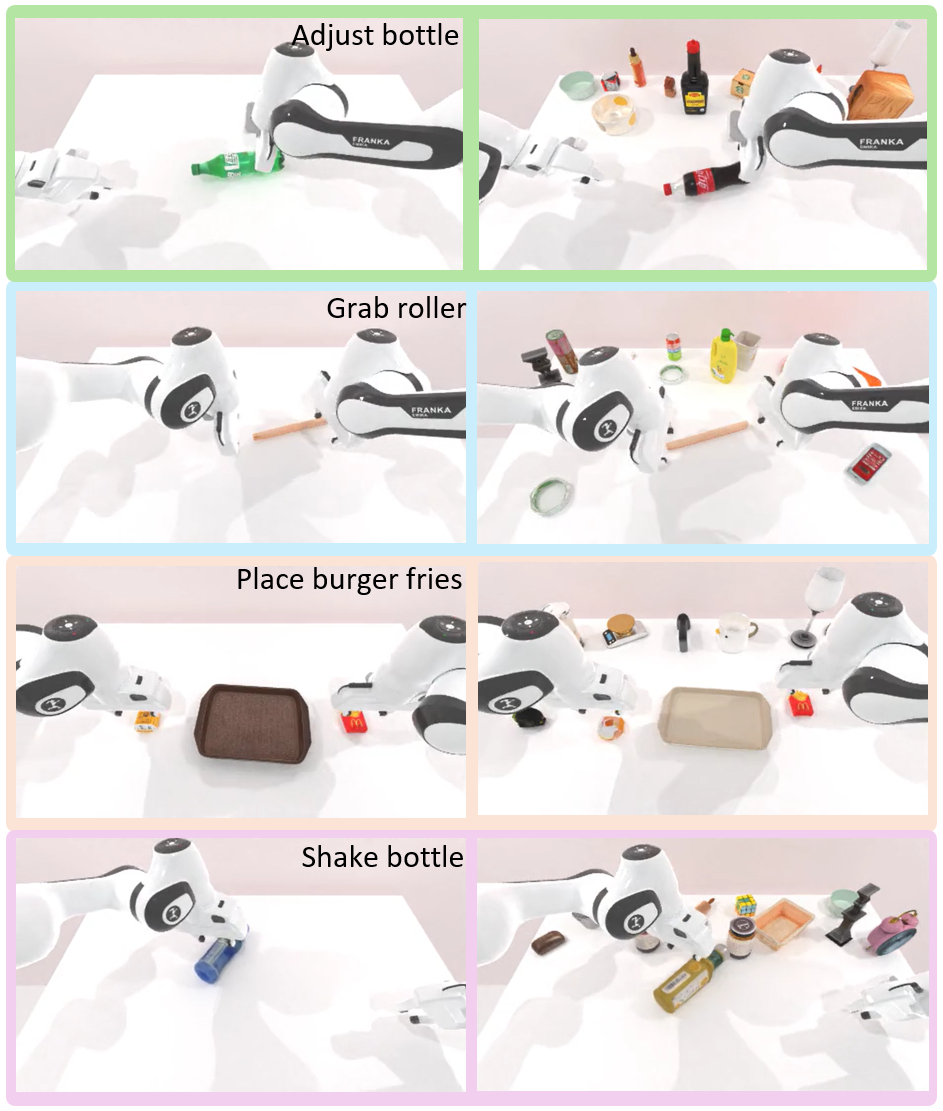} \vspace{-2mm}
    \caption{RoboTwin simulation evaluation scenes for Adjust Bottle, Grab Roller, Place Burger Fries, and Shake Bottle (top to bottom). The left column shows the ID scenes containing only task-relevant objects, while the right column shows the OOD scenes with additional visual distractors.}
    \label{fig:simulation_scenes}
    \vspace{-5mm}
\end{figure}

\subsection{Simulation Experiments}
\label{sec:simulation}

We first conduct controlled experiments on the four RoboTwin~\cite{Mu_2025_CVPR_robotwin} tasks shown in Fig.~\ref{fig:simulation_scenes}: Adjust Bottle, Grab Roller, Place Burger Fries, and Shake Bottle. These tasks cover grasping, object adjustment, placement, and dynamic manipulation. The clean and randomized RoboTwin scenes correspond to our ID and OOD settings, respectively.

For each task, all settings use the same 50 robot demonstrations. Masquerade and ours augment these demonstrations with robotized versions of the same human videos, allowing us to isolate the effect of editing quality while keeping the amount and source of training data fixed. We evaluate Diffusion Policy (DP)~\cite{chi2024diffusionpolicy} with ViT~\cite{dosovitskiy2020vit} and Swin Transformer~\cite{liu2021Swin} visual encoders, denoted DP-ViT and DP-Swin, to test whether the observed trends hold across visual representations.

We report task success rates over 100 rollouts for every policy, task, and scene. Evaluation conditions and initial-state distributions are matched across the three training settings within each scene.

As shown in Tab.~\ref{tab:robotwin_results}, both co-training methods improve average ID and OOD success over the robot-only baseline for both encoders, confirming that robotized human videos provide useful additional supervision. Our method performs best on average, reaching 68.00\%/45.75\% ID/OOD success with DP-ViT and 87.25\%/59.25\% with DP-Swin. Relative to Masquerade, the average OOD gains are 22.25 and 15.50 percentage points, whereas the corresponding ID gains are only 2.50 and 2.00 points. The larger OOD improvement suggests that more physically consistent demonstrations are especially useful when the visual context differs from the robot training scene.

The gains remain task-dependent. Our method improves OOD success over Masquerade on all four tasks with DP-ViT and on three of four tasks with DP-Swin. For DP-Swin, the OOD gains on Adjust Bottle, Grab Roller, and Place Burger Fries are 27, 31, and 19 points, respectively, but Shake Bottle falls from 36\% to 21\%. DP-ViT also drops from 85\% to 75\% ID success on Grab Roller despite improving OOD success from 12\% to 33\%. Thus, the average gains generalize across both encoders, but not uniformly across tasks and scene conditions.

\begin{table*}[t]
    \centering
    \caption{Ablation results on RoboTwin using DP-ViT. All variants are trained in the ID scene. ID and OOD denote evaluation without and with visual distractors, respectively. Success rates (\%) are measured over 100 rollouts, and averages are computed across the four tasks.}
    \label{tab:ablation_results}
    \small
    \setlength{\tabcolsep}{3.2pt}
    \begin{tabular}{l*{5}{cc}}
        \toprule
        \multirow{2}{*}{Variant}
        & \multicolumn{2}{c}{Adjust Bottle}
        & \multicolumn{2}{c}{Grab Roller}
        & \multicolumn{2}{c}{Place Burger Fries}
        & \multicolumn{2}{c}{Shake Bottle}
        & \multicolumn{2}{c}{Average} \\
        \cmidrule(lr){2-3} \cmidrule(lr){4-5} \cmidrule(lr){6-7}
        \cmidrule(lr){8-9} \cmidrule(lr){10-11}
        & ID & OOD & ID & OOD & ID & OOD & ID & OOD & ID & OOD \\
        \midrule
        Ours                         & 82 & 70 & 75 & 33 & 46 & 43 & 69 & 37 & 68.00 & 45.75 \\
        w/o Contact Reconstruction   & 83 & 41 & 84 & 24 & 47 & 26 & 69 & 30 & 70.75 & 30.25 \\
        w/o Depth-Based Occlusion    & 83 & 65 & 83 & 13 & 46 & 36 & 53 & 29 & 66.25 & 35.75 \\
        \bottomrule
    \end{tabular}
\end{table*}

\subsection{Ablation Studies}
\label{sec:ablation}

We ablate the two main components of our editing pipeline: interaction-aware contact reconstruction and depth-based robot occlusion estimation. To keep the comparison controlled and computationally efficient, all ablations use DP-ViT with the same source videos, robot demonstrations, architecture, and training configuration. Each variant removes one component while retaining the other and follows the ID/OOD protocol in Sec.~\ref{sec:simulation}.

\textbf{w/o Contact Reconstruction.}
We replace our contact reconstruction (Sec.~\ref{sec:contact_reconstruction}) with the original Masquerade grasp representation while retaining the reconstructed hand motion and depth-based occlusion estimation. This variant isolates the contribution of contact-guided grasp retargeting.

\textbf{w/o Depth-Based Occlusion.}
We composite the rendered robot directly onto the hand-inpainted frame, without comparing the inpainted-scene and rendered-robot depths as in Sec.~\ref{sec:depth_occlusion}. Contact reconstruction and retargeting remain unchanged, isolating the contribution of depth-aware compositing.

As shown in Tab.~\ref{tab:ablation_results}, removing contact reconstruction reduces average OOD success from 45.75\% to 30.25\%, despite increasing ID success from 68.00\% to 70.75\%. Removing depth-based occlusion lowers average success to 66.25\%/35.75\% in ID/OOD scenes. Both components therefore improve OOD generalization, with contact reconstruction providing the larger contribution and depth-aware compositing providing a complementary gain.

\subsection{Real-World Experiments}
\label{sec:real_world}

\begin{figure}[t]
    \centering
    \includegraphics[width=\columnwidth]{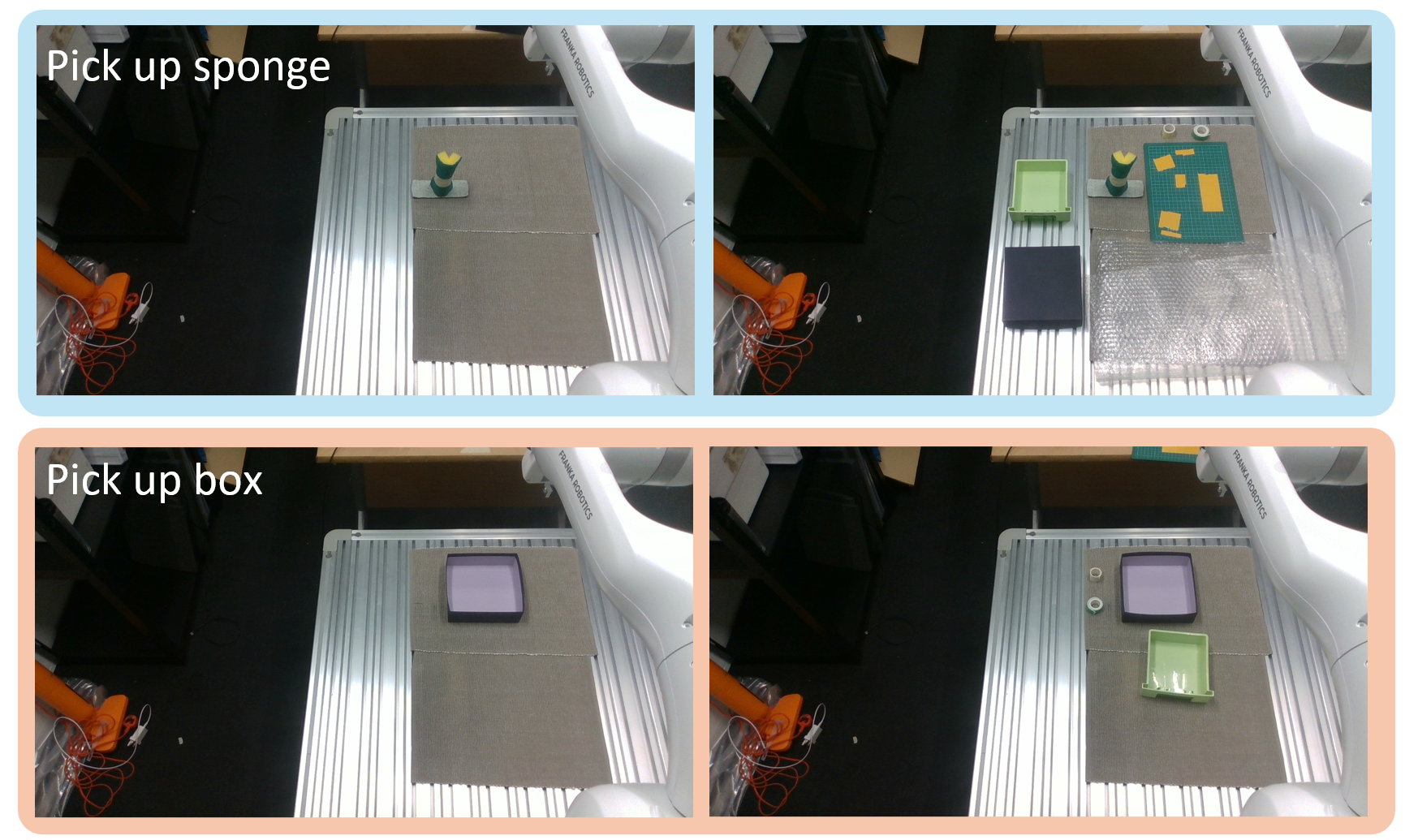}
    \vspace{-2mm}
    \caption{Real-world evaluation scenes for Pick Up Sponge (top) and Pick Up Box (bottom). The left column shows the ID scenes containing only task-relevant objects, while the right column shows the OOD scenes with additional visual distractors.}
    \label{fig:real_world_scenes}
    \vspace{-5mm}
\end{figure}

We further evaluate all three settings on a physical robot setup comprising a Franka Research 3 (FR3) robot and an Intel RealSense D435if camera, using the two tasks shown in Fig.~\ref{fig:real_world_scenes}. Pick Up Box requires a downward grasp and precise depth alignment, whereas Pick Up Sponge uses a horizontal grasp. Each task uses the same 50 real-robot demonstrations; Masquerade and ours additionally use robotized versions of the same egocentric human videos. Both co-training methods use identical human and robot data.

Following the scene definitions in Sec.~\ref{sec:exp_setup}, we evaluate each policy in both ID and OOD scenes. We conduct 10 independent trials for every combination of training setting, task, and scene and report the number of successful trials.

As shown in Tab.~\ref{tab:real_results}, co-training is most effective for Pick Up Sponge. Our method achieves 9/10 ID and 8/10 OOD successes, compared with 8/10 and 5/10 for the baseline and 9/10 and 7/10 for Masquerade. The three-point OOD improvement over robot-only training indicates greater robustness to the added visual distractors.

Pick Up Box remains difficult for all methods: ours achieves 4/10 ID and 2/10 OOD successes. Unlike the horizontal sponge grasp, the downward box grasp demands precise depth estimation for gripper alignment. Because the setup uses only a single egocentric camera, limited depth cues constrain performance even with co-training. Overall, our method improves real-world OOD performance for the horizontal sponge grasp, but the box results show that better demonstrations cannot fully compensate for insufficient geometric observability.

Overall, the real-world results support the value of physically consistent human-video co-training, particularly for improving robustness to visual distractors. However, its benefit depends on the perceptual demands of the task: better edited demonstrations help when the required geometry is observable, but cannot resolve ambiguities caused by insufficient depth information. These findings highlight that scalable human-video supervision and reliable geometric perception are complementary requirements for robust real-world manipulation.

\begin{table}[t]
    \centering
    \caption{Real-world task success over 10 trials per task and evaluation scene. ID and OOD denote evaluation without and with visual distractors, respectively. Bold indicates the best result in each setting.}
    \label{tab:real_results}
    \small
    \setlength{\tabcolsep}{8pt}
    \begin{tabular}{lcccc}
        \toprule
        \multirow{2}{*}{Training data}
        & \multicolumn{2}{c}{Pick Up Box}
        & \multicolumn{2}{c}{Pick Up Sponge} \\
        \cmidrule(lr){2-3} \cmidrule(lr){4-5}
        & ID & OOD & ID & OOD \\
        \midrule
        Baseline   & \textbf{4}/10 & 1/10          & 8/10          & 5/10 \\
        Masquerade & 3/10          & \textbf{2}/10 & \textbf{9}/10 & 7/10 \\
        Ours       & \textbf{4}/10 & \textbf{2}/10 & \textbf{9}/10 & \textbf{8}/10 \\
        \bottomrule
    \end{tabular}
\end{table}
\section{Conclusion}
\label{sec:conclusion}
In this work, we introduced an interaction-aware human-to-robot data-editing pipeline that converts egocentric human videos into robot-compatible demonstrations while preserving physical consistency. Our method combines contact-based grasp reconstruction for stable parallel-jaw retargeting with depth-aware compositing for correct robot--object visibility. Across simulation and real-world experiments, the resulting demonstrations improve policy learning over robot-only training and Masquerade-based editing, with particularly clear gains under out-of-distribution scene variation. These results suggest that the value of human-video supervision depends not only on scale, but also on how faithfully the original interaction is preserved during embodiment conversion. By improving both grasp geometry and scene consistency, our approach provides a practical path toward higher-quality robot-learning supervision from abundant human videos.

More broadly, our findings highlight that the value of human-video supervision depends not only on data scale, but also on how faithfully the original interaction is preserved during embodiment conversion. Contact reconstruction improves the geometric validity of grasp retargeting, while depth-aware compositing preserves the visual evidence of object interaction. By addressing both, our method provides a practical path toward converting abundant human videos into higher-quality robot-learning supervision. Future work will extend this framework to dexterous hands, more complex object geometries, and larger collections of unconstrained human videos.

\bibliographystyle{ieeetr}
\bibliography{ref}

\end{document}